\documentclass{article}

\usepackage[final]{neurips_2024}

\usepackage[utf8]{inputenc} 
\usepackage[T1]{fontenc}    
\usepackage{hyperref}       
\usepackage{url}            
\usepackage{amsfonts}       
\usepackage{nicefrac}       
\usepackage{microtype}      
\usepackage{xcolor}         

\usepackage{graphicx}

\usepackage{subcaption}     
\usepackage{booktabs}       
\usepackage{tabularx}
\usepackage{colortbl}
\usepackage{adjustbox}
\usepackage{mparhack}

\usepackage{amsmath, dsfont, amssymb}
\usepackage{algorithm}
\usepackage[noend]{algpseudocode}

\usepackage{wrapfig}
\usepackage{mathrsfs}
\usepackage{rotating}
\usepackage{cleveref}

\definecolor{Gray}{gray}{0.9}
\definecolor{LightCyan}{rgb}{0.88,1,1}
\definecolor{golden}{rgb}{255, 215, 0}

\newcolumntype{R}[2]{%
    >{\adjustbox{angle=#1,lap=\width-(#2)}\bgroup}%
    l%
    <{\egroup}%
}

\newtheorem{definition}{Definition}

\newcommand{\methodname}{HeLU}

\title{Hysteresis Activation Function for Efficient Inference}

\author{%
  Moshe Kimhi\thanks{Equal contribution.}\\
  Technion - Israel Institute of Technology\\
  Department of Computer Science\\
\texttt{moshekimhi@cs.technion.ac.il} \\
  \And
  Idan Kashani$^{\ast}$ \\
    Technion - Israel Institute of Technology\\
  Department of Computer Science\\
  \texttt{idan-kashani@cs.technion.ac.il} \\
    \And
  Avi Mendelson\\
    Technion - Israel Institute of Technology\\
  Department of Computer Science\\
    \And
  Chaim Baskin\\
  Ben-Gurion University of the Negev\\
  School of Electrical and Computer Engineering\\
}

\begin{document}

\maketitle

\begin{abstract}

The widely used ReLU is favored for its hardware efficiency, {as the implementation at inference is a one bit sign case,} yet suffers from issues such as the ``dying ReLU'' problem, where during training, neurons fail to activate and constantly remain at zero, as highlighted by Lu et al.~\citep{lu2018collapse}. Traditional approaches to mitigate this issue often introduce more complex and less hardware-friendly activation functions. In this work, we propose a Hysteresis Rectified Linear Unit (HeLU), an efficient activation function designed to address the ``dying ReLU'' problem with minimal complexity. Unlike traditional activation functions with fixed thresholds for training and inference, HeLU employs a variable threshold that refines the backpropagation. This refined mechanism allows simpler activation functions to achieve competitive performance comparable to their more complex counterparts without introducing unnecessary complexity or requiring inductive biases. Empirical evaluations demonstrate that HeLU enhances model generalization across diverse datasets, offering a promising solution for efficient and effective inference suitable for a wide range of neural network architectures.
\end{abstract}    
\section{Introduction}
\label{sec:intro}

The ``dying ReLU'' problem, as observed with the ReLU activation function, occurs because ReLU outputs zero for negative inputs. This can lead to neurons becoming inactive during training, thereby halting their learning capability~\citep{lu2018collapse}. This phenomenon significantly impacts the network's ability to adapt and generalize effectively. Various alternatives to ReLU have been proposed \citep{he2015delving, hendrycks2023gaussian, misra2019mish, ramach2017searching, shazeer2020glu}, which allow small negative values to persist, and thus avoid ``dying ReLU'' during training. However, these modifications often come at the cost of less efficient inference time.

\begin{figure}[htbp]
    \begin{center}
    \includegraphics[width=0.8\textwidth]{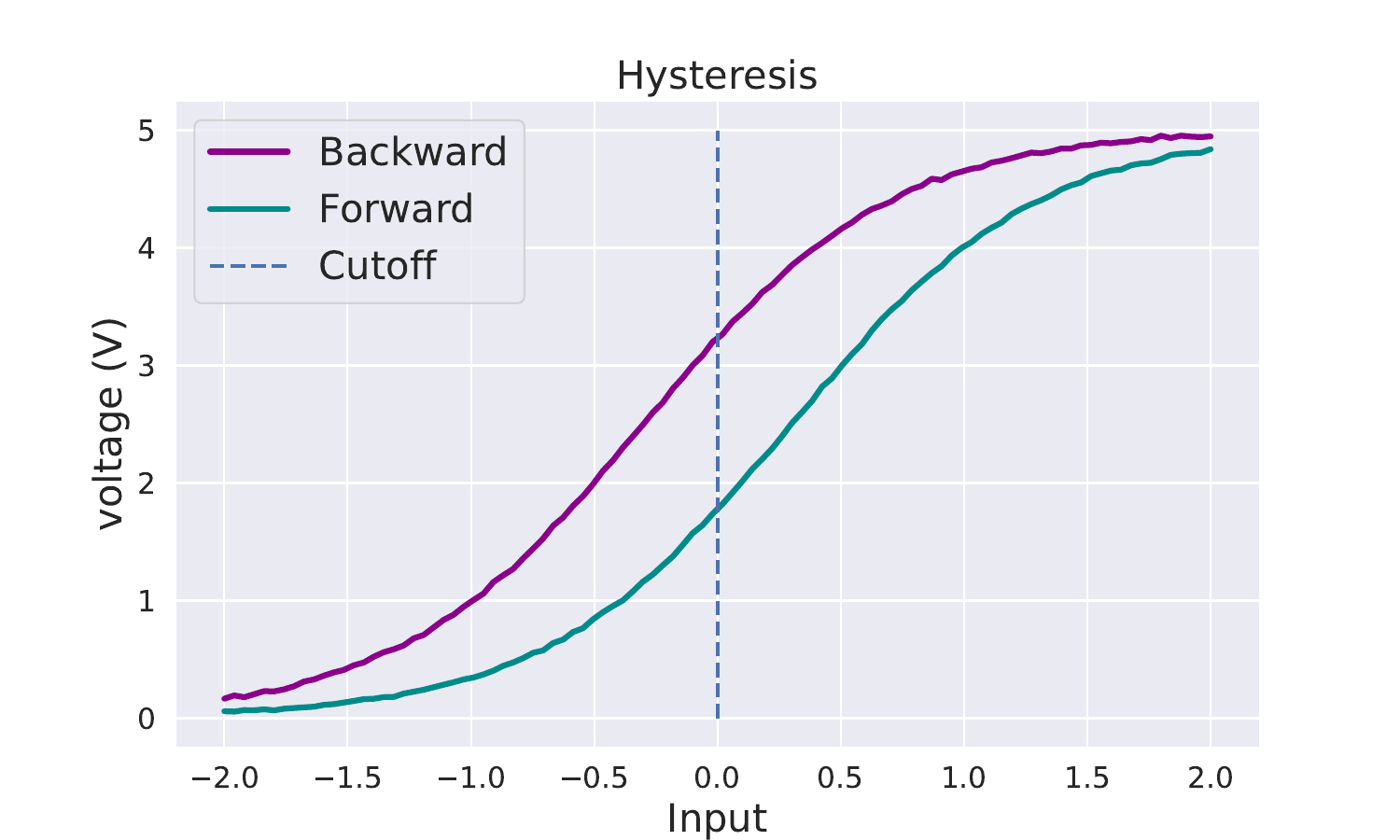}
    \caption{The threshold transition of bit voltage between 0 and 1 (forward) and 1 and 0 (backward) inspired the concept of using different thresholds for activation functions in the forward and backward passes of the network.}
    \label{fig:qualitive_output}
    \end{center}
\end{figure}


This inspired us to explore the use of robust mechanisms, with zero computational cost, by embracing Hysteresis \cref{fig:qualitive_output}.
Hysteresis, derived from the Greek word meaning ``lagging behind'', describes a fundamental property observed in various physical and engineering systems, including computer architecture. In computational contexts, hysteresis refers to the persistence of a state or behavior even after the input conditions that triggered it have changed or been removed. In computer architecture, hysteresis is prominently observed in memory elements such as flip-flops and in signal processing circuits. It plays a crucial role in ensuring proper timing and synchronization, preventing unintended state changes due to noise or transient signals. Engineers utilize hysteresis to design robust and dependable circuits capable of maintaining stable outputs amidst varying input conditions, thereby enhancing overall system performance and reliability.

\begin{figure}[htbp]
    \begin{center}
    \includegraphics[width=0.8\textwidth]{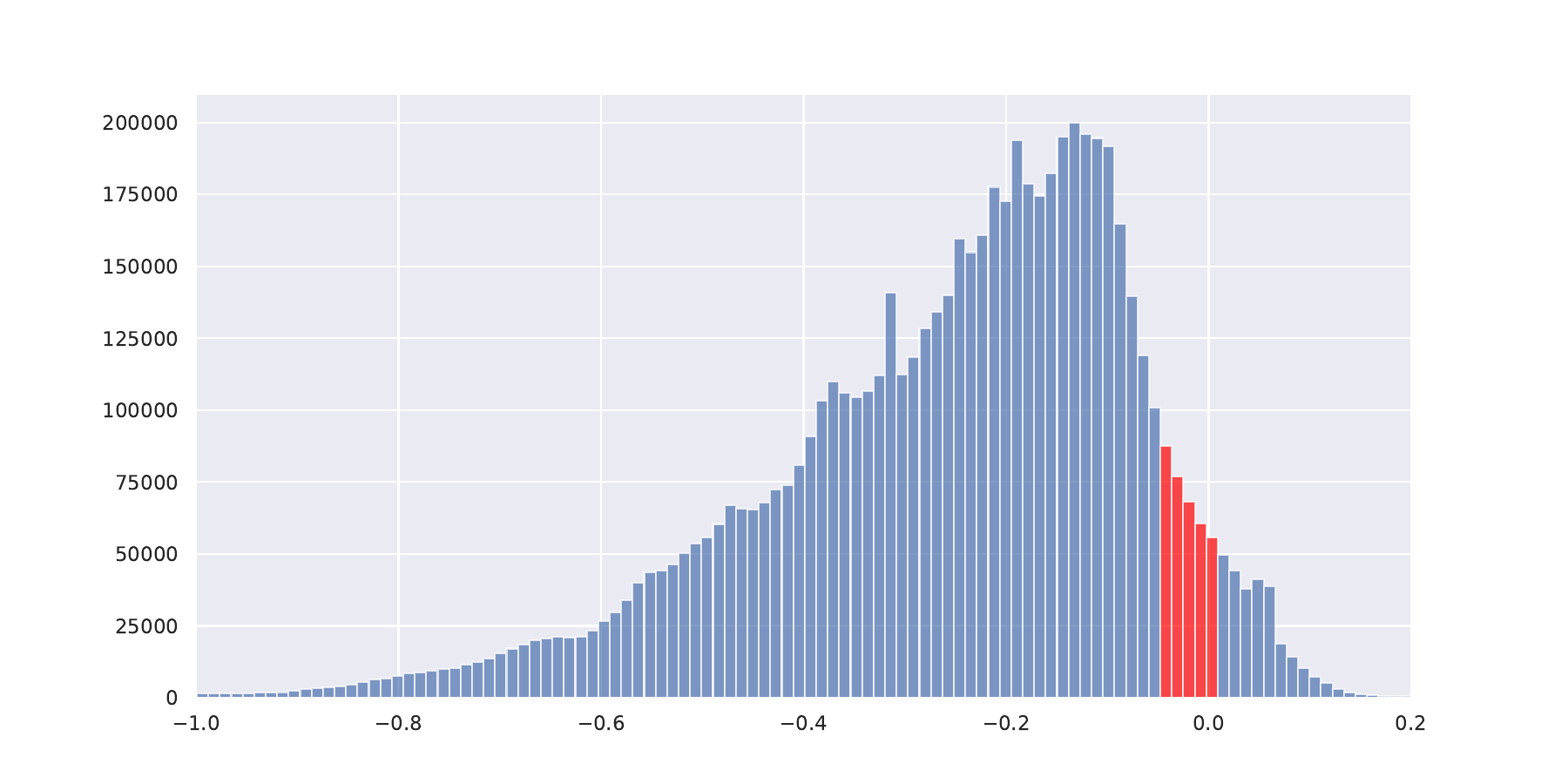}
    \caption{Weight distribution of all pre-activations of Wide ResNet 40-4. In the image classification task with over parametric network, we see that majority of the pre-activation features are negative. In red we mark the pre-activations that we allow regret with \methodname{}.}
    \label{fig/dist}
    \end{center}
\end{figure}

By introducing hysteresis, which adjusts the activation threshold, used during backpropagation, the proposed \methodname{} mitigates the ``dying ReLU'' problem without incurring additional computational costs during inference and only negligible costs during training. This approach enables neurons to remain active even for inputs that would otherwise result in zero activation in traditional ReLU networks (marked in red in \cref{fig/dist}). By maintaining non-zero gradients and promoting continuous learning in potentially dormant neurons, Hysteresis ReLU offers a promising direction for achieving attractive inference properties in lean networks with comparable performance. For instance, we observed a performance gain of +2.96 on CIFAR10, +2.19 on CIFAR100 and +1.23 over Imagenette in the computer vision domain, and +0.51 on the GLUE benchmark across 8 datasets, all compared to the simple ReLU.

\label{sec:related}
\section{Related Work}

To increase throughput and reduce energy consumption in growing DNN models for deployment on edge devices, common methods include pruning and quantization. Pruning techniques remove less significant weights or features, requiring sophisticated re-routing in hardware to achieve actual benefits~\citep{diao2023pruning, pasandi2020modeling}. Quantization, another prominent approach, reduces the precision of weights and activations to fixed precisions such as INT8 or INT4, or even binary representation~\citep{banner2019posttraining, math10214107}, and mixed-precision~\citep{micikevicius2018mixed, rakka2022mixedprecision, math12121810}. This reduces the memory footprint and enables faster matrix multiplications, which are the most common operations in DNNs.

We argue that for accelerated inference, focusing on the most fundamental building blocks, such as activation functions, can reveal further improvements in throughput. These improvements are particularly significant for more efficient models. 

Traditional methods may overlook the computational efficiency of simpler components like activation functions, which, when optimized, can lead to substantial performance gains. Specifically, the ReLU activation function, due to its simplicity and computational efficiency, demonstrates the potential for enhanced throughput without the need for additional multiplications. Here we explore some alternative activation functions that does not benefit from this property.

\subsection{Activation Functions}
A deep neural network (DNN) is a parametric approximation of a function, constructed by repeatedly applying a linear transformation to the input followed by a non-linearity. These non-linear functions, known as activation functions, are crucial for enabling the network to learn complex mappings. Over time, the selection of activation functions has been influenced by both engineering and scientific considerations~\cite{dubey2022activation}. Enhancing the power of activation functions can improve the performance of models like transformers, often by incorporating additional operations such as products and linear projections, as seen in Gated Linear Units (GLUs)~\citep{shazeer2020glu}.

\textbf{Sigmoid} $(f(x)=\sigma(x)=\frac{1}{1+e^{-x}})$, which was designed to mimic the firing rate of biological neurons. However, the Sigmoid function is susceptible to the ``vanishing gradients'' problem, where repeated multiplications during backpropagation result in gradients that approach zero, hindering the training process~\citep{ven2021regularization}.

\textbf{ReLU} ($f(x)=\max\{0,x\}$)~\citep{ReLU} activation function is the most prominent one in DNNs for many years, due to its empirical success. 
It is simple to implement in hardware, as signed representations simply look at the sign bit (MSB) and pass forward the number itself, or zero if negative. This means it does not require any intermediate calculations or additional parameters. In other words, the Arithmetic logic unit (ALU) of a processor does not need to execute any mathematical unit, unlike other cased activation functions. ReLU does not suffer from vanishing gradient, but from ``dying ReLU'' phenomenon, where a neuron might be deactivated during training producing a constant output of 0, and gradient updates would not change that.

In the recent years, \textbf{GELU} ($f(x)=x\Phi(x)$)~\citep{hendrycks2023gaussian} where $\Phi(x)$ is the Gaussian Error function, has proven great success in many transformer models and has become a standard to use. Typically used with approximation
$ \frac{x}{2} [ 1+Tanh (\sqrt{\frac{2}{\pi}}(x+0.044715x^3)] $, that even if faster, still require computation of the Tanh function. Other similar functions such as \textbf{Swish} ($f(x)=x\sigma(x)$)~\citep{ramach2017searching}, have become common in vision models.
GELU is inspired by stochastic regularization and was formulated with a probabilistic approach of gating. It views the activation of a neuron and the following dropout as a single component, that is produced by the neuron.
One thing to note about GELU is its curve, where its derivative is negative. This is in contrary to the former known activation functions, that are monotonic, and serves as a mean of regularization over the gradients, encouraging optimization to be more robust to sub-optimums, as local minima and saddle points.
We refer the reader to LLMCompas~\citep{zhang2023hardware}, that analyze the hardware complexity and utilization of LLMs, showing that for GPT3, the GELU operations cause 400-500 G elements per second on NVIDIA A100. Moreover, the analysis shows that specifically for GPU usage, GELU can cause around $6\%$ of the latency of GPT3, where quantized models become more popular, increasing this number significantly.

A recent trend of trainable activation functions gains popularity, with the notable Kolmogorov-Arnold Network (\textbf{KAN}) \citep{liu2024kan}. However empirical evident shown learnable activation functions are extremely inefficient, as they can take up to 10 times longer to train, and do not have any inference utilization grantees~\citep{dubey2022activation}, they left out of the scope of this work. 

Despite the advancements in activation functions, many of these alternatives do not benefit from the computational simplicity of ReLU, which merely requires a single-bit condition check and no multiplications. This makes ReLU particularly attractive for applications requiring fast and efficient inference, highlighting the trade-off between the advanced capabilities of newer functions and the hardware efficiency of ReLU, such as this work.





\section{\methodname{}}
\label{sec:method}
\begin{figure}
    \begin{center}
    \includegraphics[width=0.85\linewidth]{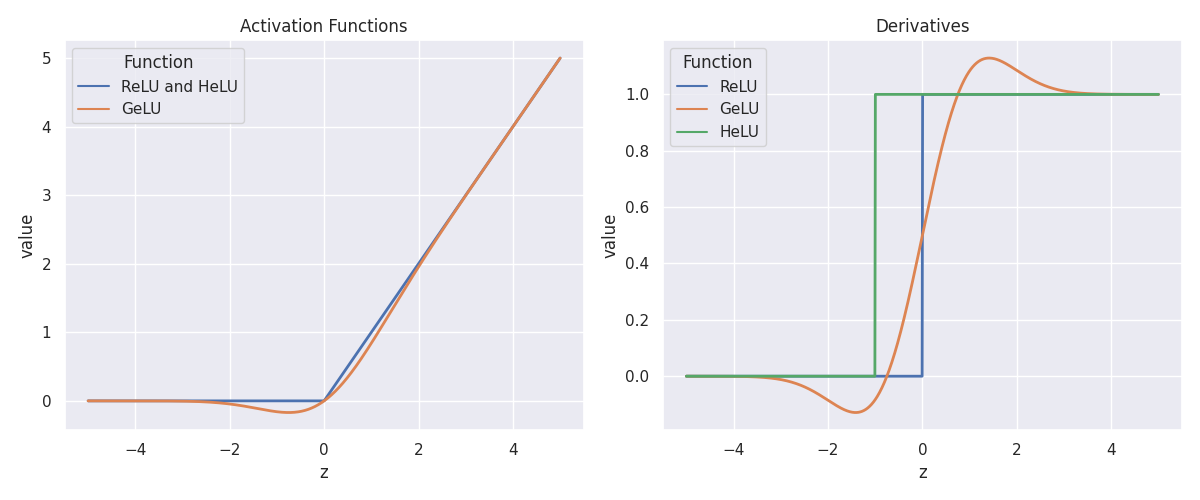}
    \caption{ReLU, GELU and \methodname{} functions and their derivatives}
    \label{fig:funcPlots}
    \end{center}
\end{figure}

Our formulation leverage the robustness of hysteresis, that preforms regularization by setting different threshold for the forward and backward steps on neurons of network, during training, as seen in \Cref{fig:funcPlots}.
\methodname{} is defined exactly like ReLU, for the forward pass, but for the backpropagation, it is shifted backwards by a hyperparameter, $\alpha \in \mathbb{R}$.

We define the Hysteresis Rectifier Linear Unit (\methodname{}) as:
\begin{definition}[\methodname{}]
    $\methodname{}_{\alpha}(x) = ReLU(x) = \max\{0, x\}$
\end{definition}

Whereas, we override the autograd derivative that is used during backpropagation, and use instead the following term, lightly abusing the notation:

\begin{definition}[\methodname{} Modified Derivative]
\(\frac{d}{dx}\) $\methodname{}_{\alpha}(x) = $ \[
        \left\{\begin{array}{lr}
        0 & x \leq -\alpha\\
        1 & -\alpha < x\\
        \end{array}\right\}
  \] \end{definition}

Shifting back the derivative, essentially, refines the trigger of ``dying ReLU'', by demanding a greater step to turn off the neuron.
For convenience, we added a simple implementation of \methodname{} in Torch-like pseudo code in \Cref{algo}.

\begin{algorithm}
\caption{\methodname{} Function in PyTorch Style}
\label{algo}
\begin{algorithmic}[1]
\State \textbf{class} \texttt{\methodname{}Function(torch.autograd.Function)}

\State \texttt{@Staticmethod}
\State \textbf{def} \texttt{forward(ctx, z, alpha)}
\State \quad \texttt{ReLU} $\gets$ \texttt{torch.where(z > 0, z, 0)}
\State \quad \texttt{ctx.save\_for\_backward(z)}
\State \quad \texttt{ctx.alpha} $\gets$ \texttt{alpha}
\State \quad \textbf{return} \texttt{ReLU}

\State \texttt{@Staticmethod}
\State \textbf{def} \texttt{backward(ctx, grad\_output)}
\State \quad \texttt{z,} $\gets$ \texttt{ctx.saved\_tensors}
\State \quad \texttt{alpha} $\gets$ \texttt{ctx.alpha}
\State \quad \texttt{grad\_positive} $\gets$ \texttt{torch.ones\_like(z)}
\State \quad \texttt{grad\_\methodname{}} $\gets$ \texttt{torch.where(z > -alpha, grad\_positive, 0)}
\State \quad \textbf{return} \texttt{grad\_\methodname{} * \texttt{grad\_output}}

\end{algorithmic}
\end{algorithm}

\section{Experiments}
\label{sec:experiments}
We evaluate the performance of \methodname{} along with common counterpart activations (GELU, ELU, and ReLU) across two prominent domains: computer vision and natural language processing, specifically for discriminative tasks. Our results indicate that \methodname{} performs comparably to GELU and outperforms ReLU while retaining the computational efficiency of ReLU. Consequently, our experiments emphasize the accelerated paradigm of the aforementioned tasks, as described in this section.

\subsection{Image Classification}
\label{sec:CV}
We conducted experiments on CIFAR10 and CIFAR100 datasets \citep{krizhevsky2009cifar}, as well as on a subset of ImageNet containing 10 common classes, known as Imagenette \citep{imagewang}. Consistent with \citep{hendrycks2023gaussian}, all image classification experiments employed the 40-4 Wide ResNet architecture \citep{zagoruyko2016wide}. For the CIFAR datasets, we trained for 100 epochs with a learning rate of 0.01, utilizing the momentum SGD optimizer with standard horizontal flip and crop augmentations. For Imagenette, we trained for 50 epochs with the same optimization algorithm but without augmentations. As shown in Table \ref{table:cv}, \methodname{} demonstrates superior performance compared to the non-shifted threshold ReLU function, outperforming other methods or being second only to GELU in certain cases, such as CIFAR.

\begin{table}[ht]
\centering
\caption{
Image classification Accuracy of Wide ResNet 40-4 with various activations.
}
\label{table:cv}
\begin{tabular}{l|c|c|c}
\toprule
 \rowcolor{Gray} Activation Function & CIFAR10 & CIFAR100 & Imagenette \\
\midrule
\midrule
ReLU & $92.84 \pm 0.34$ &$75.31 \pm 0.94$& $76.76 \pm 1.50$ \\
ELU  & $93.59 \pm 0.75$ &$71.85 \pm 1.77$&  $76.67 \pm 2.00$\\
GELU & $\textbf{96.11} \pm 0.51$  & $\textbf{79.26} \pm 0.95$ & $\textbf{78.45} \pm 2.51$\\   
\hline
\methodname{} ($\alpha=0.001$) & $\underline{95.80} \pm 0.63$ &$\underline{77.51} \pm 1.20$& $\underline{77.99} \pm 1.21$\\ 
\methodname{} ($\alpha=0.05$) & $95.31 \pm 0.73$ &$75.62 \pm 1.47$& $76.67 \pm 2.35$\\  
\methodname{} ($\alpha=0.01$) & $95.35 \pm 0.94$ & $75.84 \pm 1.95$& $75.73 \pm 2.85$\\  
\methodname{} ($\alpha=0.1$) & $95.10 \pm 0.77$ & $68.75 \pm 1.45$& $72.32\pm 4.52$\\  
\midrule
\bottomrule
\end{tabular}
\end{table}

\subsection{General Language Understanding Evaluation }
\label{sec:nlp}


\textbf{Model Architecture and Training}

Our experiments were conducted using \textit{Cramming}~\citep{geiping2022cramming}, a highly optimized training framework of \textit{BERT-base-uncased}~\citep{bert} model, including many techniques to boost the training and the inference of models, like automatic mixed precision (AMP)~\citep{micikevicius2018mixedprecisiontraining}, flash attention~\citep{dao2022flashattention}, torch compile~\citep{Ansel2024PyTorch2F}.
All the added optimizations are orthogonal to \methodname{}, and we believe one might like to use them in parallel with \methodname{}, to speed up inference.
The training objective was also slightly modified, compared to the original \textit{BERT}, including only masking loss, and not next sentence prediction, as shown in \textit{RoBERTA}~\citep{liu2019roberta}. 
The pretraining phase took 3327000 steps, out of them, 1\% for warmup.

Hyperparameters for BERT pretraining were selected by ablation and are specified in \Cref{tab:bert_pretraining_hyperparams}. At first, we performed a hyperparameter search on the GELU baseline, mainly to find the appropriate learning rate.
After that, we used the selected hyperparameters to train another 3 \methodname{} models, where all GELU activations were replaced with \methodname{} with the following alphas: ${0.05, 1, 2}$, and one ReLU model with ReLU activations.
We used wikipedia-bookcorpus as \textbf{Pretraining Data}, similar to the original paper of \textit{BERT}. 

\noindent \textbf{Evaluation Benchmarks}

After pretraining, we evaluated the models on 8 out of the 9 natural language understanding tasks included in GLUE~\citep{wang2019glue}. Every model was fine-tuned for each of the tasks, using the \textit{Cramming} code base.
The results reflected from Table \ref{tab:glue_comparison} shows an improvement using \methodname{} $\alpha=0.05$ over ReLU. Inline with the Vision results, GELU still achieves more accurate results on most datasets. Note that \methodname{} aim to bridge the gap between ReLU and GELU without memory and compute additional cost.

Another experiment conducted on QBERT \citep{shen2019qbert}, a post-training quantization of the BERT model without any post-training calibration, as shown in Table \ref{tab:glue_comparison2}, highlights the gap in performance between QBERT and BERT. The results indicate that the tested activation functions suffer a small degradation in performance when quantized to 8-bit, whereas \methodname{} shows a mild improvement.

\begin{table}
\centering
\caption{
Comparison of different activation functions on GLUE tasks (Larger is better).
}
\label{tab:glue_comparison}
\resizebox{1.\linewidth}{!}{
\begin{tabular}
{p{3cm}p{1.5cm}p{1.5cm}p{1cm}p{1.5cm}p{1.5cm}p{1.5cm}p{1cm}p{1.5cm}p{1.5cm}p{1cm}}
\toprule
\rowcolor{Gray} Name & \multicolumn{2}{c}{MNLI Accuracy} & QQP & QNLI & SST-2 & STS-B & MRPC & RTE & WNLI & Avg \\
\rowcolor{Gray} & (m) & (mm) & F1 & Acc & Acc & Spearman & F1 & Acc & Acc & \\
\midrule
\midrule

GELU & 81.9 & 82.52 & \textbf{86.6} & \textbf{89.3} & \textbf{90.8}  & \textbf{82.4} & 82.0 & 53.1 & 56.3 & \textbf{78.32} \\   

ReLU & \textbf{82.7} & \textbf{82.9} & 85.7 & 87.8 & 90.3  & 74.3 & \textbf{83.2} & 55.2 & 47.9 & 76.67 \\ 

\methodname{} $\alpha=0.05$ & \underline{82.0} & \underline{82.59} & \underline{86.0} & \textbf{89.3} & \underline{90.6}  & \underline{76.9} & 81.6 & \textbf{56.3} & \textbf{57.7} & \underline{78.02} \\ 

\methodname{} $\alpha=1$ & 81.4 & 81.62 & 84.7 & 88.0 & 86.5 & 18.2 & 81.3 & 54.2 & \textbf{57.7} & 70.4 \\  

\methodname{} $\alpha=2$ & 32.7 & 32.95 & 70.0 & 78.0 & 50.9 & 13.2 & 81.2 & 54.5 & 53.5 & 51.88 \\  

\midrule
\bottomrule
\end{tabular}
}
\end{table}

\begin{table}[ht]
\centering
\caption{Comparison of different activation functions on GLUE tasks, showing the difference between QBERT-INT8 and BERT performance (Larger is better).}
\label{tab:glue_comparison2}
\resizebox{\linewidth}{!}{
\begin{tabular}{p{3cm}p{1.5cm}p{1.5cm}p{1.1cm}p{1.5cm}p{1.5cm}p{1.5cm}p{1.1cm}p{1.5cm}p{1.5cm}p{1.1cm}}
\toprule
\rowcolor{Gray} Name & \multicolumn{2}{c}{MNLI Accuracy} & QQP & QNLI & SST-2 & STS-B & MRPC & RTE & WNLI & Avg \\
\rowcolor{Gray} & (m) & (mm) & F1 & Acc & Acc & Spearman & F1 & Acc & Acc & \\
\midrule
\midrule
ReLU & -0.4589 & -0.2238 & \textbf{0.0726} & \textbf{0.0915} & \textbf{0.4587} & 0.0284 & \textbf{0.1316} & 0.3610 & -2.8169 & -0.2617 \\
GELU & -1.4978 & -2.0952 & -0.1234 & 0.0732 & 0.3440 & -0.0123 & 0.0535 & \textbf{1.0830} & 0.0000 & -0.2417 \\
\methodname{} $\alpha=0.05$ & \textbf{0.1822} & \textbf{1.6273} & -0.1400 & -0.0549 & 0.3440 & \textbf{0.2296} & 0.1251 & 0.7220 & \textbf{4.2254} & \textbf{0.8067} \\
\bottomrule
\end{tabular}
}
\end{table}

\begin{figure}[ht]
    \begin{center}
    \includegraphics[width=\linewidth]{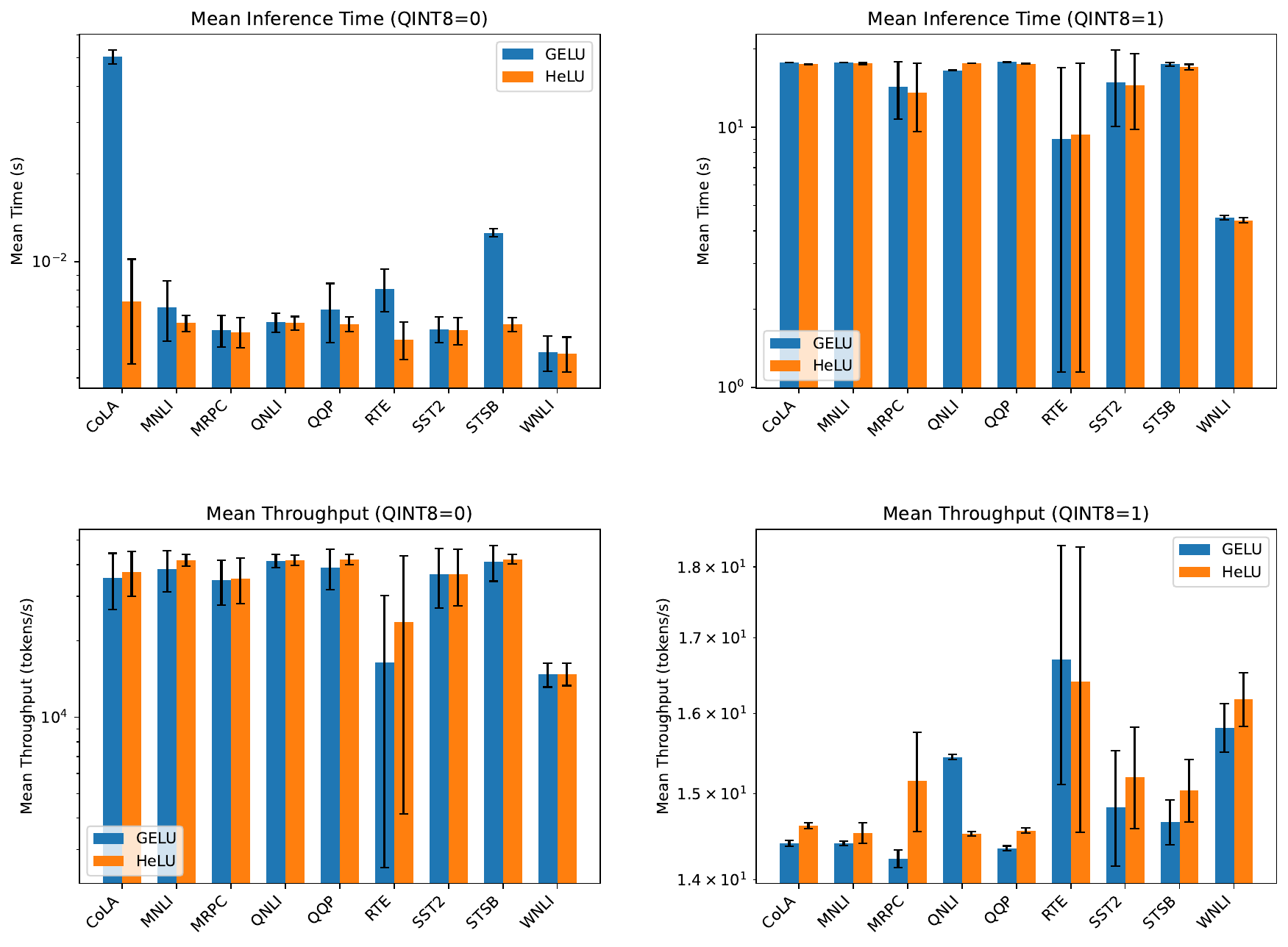}
    \caption{Comparing \methodname{} and GELU in BERT. Top figures are Inference time by dataset, using half precision and QINT8 versions. Bottom figures are Throughput using half precision and QINT8.}
    \label{fig:timing_w_quant}
    \end{center}
\end{figure}

\begin{figure}[ht]
    \begin{center}
    \includegraphics[width=0.75\linewidth]{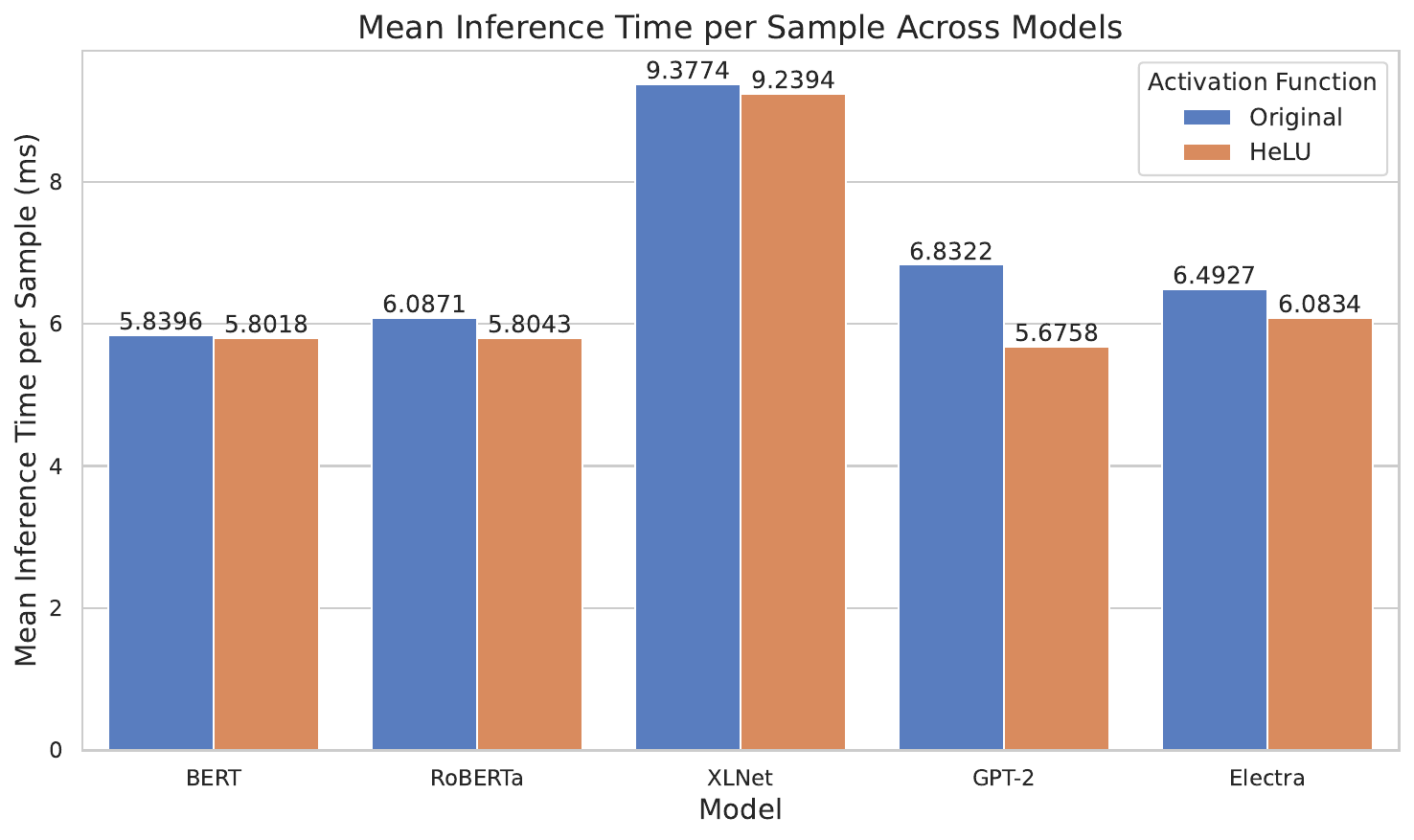}
    \caption{Analysis of various transformer architectures inference time per sample, using their original activation and \methodname{}.}
    \label{fig:models}
    \end{center}
\end{figure}

\noindent \textbf{Training Environment}

We used 3 $\times$ \textsc{NVIDIA GeForce RTX 2080 Ti}. Each GPU has memory capacity of close to 12GB. Each pretraining took us approximately 1 week.
The seed for the pretraining, fine-tuning and evaluation of all tested models is 42.

The pretraining of the GELU models was using \texttt{CUDAGraphs}~\citep{nguyen2024cudagraphs}, an optimization mechanism that allows a series of CUDA kernels to be executed as a single unit, reducing the launching overhead and achieving speedup, provided by the training framework. However, we had to omit the optimization when pretraining \methodname{} and ReLU models due to a lack of memory required for torch compilation. Thus, we consider a further optimized version of the GELU model in our comparison.








\subsection{Computational Analysis}
\label{sec:ablation}


We assess the inference time and throughput of BERT with and without QINT8 quantization (using only AMP for the latter) by running quantized models on a CPU and non-quantized models on an NVIDIA L40 GPU. The evaluation iterates over batches of 256 drawn from a subset of up to 1024 randomly selected samples from the validation set of each GLUE benchmark dataset, with measurements repeated 10 times. The results of this experiment are illustrated in \Cref{fig:timing_w_quant}. Our findings reveal a notable trend: \methodname{} achieves faster inference times, particularly in the non-quantized model, and higher throughput, especially in the quantized model, where memory constraints are more pronounced.
The reported standard deviation is calculated based on batch-wise processing rather than individual samples, reflecting the variability introduced by the random selection of samples from the validation set. Consequently, datasets with considerable variation in sequence length may exhibit high variance, though this pattern remains consistent across different non-linearity functions.

{Another experiment presented in \Cref{fig:models}, shows inference time per sample using various transformer architectures, namely BERT \cite{bert}, RoBERTa \cite{liu2019roberta}, XLNet \citep{yang2019xlnet}, GPT-2 \cite{radford2019language} and Electra \cite{clark2020electra} on single NVIDIA T4 GPU. We denote the original implementation from \cite{wolf-etal-2020-transformers} as Original, since some models like GPT-2 used a modified version of GELU. In all models we observed a reduction in inference time, where the experiment averaged over $100,000$ examples. The total timing of the examples benefit range from 121.63 seconds (GPT-2) to only 4.735 seconds (BERT). While BERT seem insignificant, we show in \Cref{fig:timing_w_quant} that in some cases, like the CoLA or STSB datasets it can be very significant.}



\begin{table}[ht]
    \centering
    \begin{tabular}{l|c}
        \toprule
        \rowcolor{Gray} \textbf{Hyperparameter} & \textbf{Value} \\
        \midrule
        \midrule
        Batch Size & 288 \\
        Gradient Clipping & 0.5 \\
        Optimizer & AdamW \\
        Learning Rate & 0.0001 \\
        Betas & [0.9, 0.98] \\
        Epsilon & 1E-12 \\
        Weight Decay & 0.01 \\
        Total Steps & 3,327,000 \\
        Warmup Steps & 33,270 \\
        Scheduler & Budget-One-Cycle \\
        \midrule
        \bottomrule
    \end{tabular}
    \caption{Pretraining Hyperparameters of BERT}
    \label{tab:bert_pretraining_hyperparams}
\end{table}

\section{Conclusion}
\label{sec:Discussion}
In this study, we advocate for using shifted activation threshold during training, \methodname{}, a novel hysteresis activation function designed to enhance the robustness of activations against the ``dying ReLU'' phenomenon. By leveraging the gradients of a shifted activation function, our approach maintains simplicity while improving performance across various domains. Our empirical results indicate that a relatively straightforward modification can yield significant performance gains, highlighting the potential of \methodname{} for universal application. This study paves the way for exploring efficient activation functions that are both computationally and environmentally sustainable.
While extremely efficient, one drawback of \methodname{} is the search for the optimal hyper-parameter; Future research should further investigate the pre-activation statistics and find an optimal value for $\alpha$.
Explore the impact on other domains such as segmentation, specifically under reduced supervision  \cite{kimhi2023semi,kimhi2025robot} or noisy labeled \cite{noisy}, could reveal additional benefits due to the noisy training process.
It is important to note that \methodname{} is intended to enhance efficiency rather than replace activation functions in systems where accuracy is paramount.


\bibliographystyle{abbrvnat}
\bibliography{main}
\end{document}